\documentclass[10pt,twocolumn,letterpaper]{article}
\usepackage{iccv}
\usepackage{times}
\usepackage{epsfig}
\usepackage{graphicx}
\usepackage{amsmath}
\usepackage{amssymb}

\newcommand{\figref}[1]{Fig.~\ref{#1}}
\newcommand{\tabref}[1]{Tab.~\ref{#1}}
\newcommand{\secref}[1]{Sec.~\ref{#1}}

\newcommand{\equref}[1]{Eqn.~(\ref{#1})}
\usepackage{booktabs}
\usepackage{multirow}
\usepackage{overpic}
\usepackage{enumitem}
\usepackage{xcolor}
\usepackage{subfig}
\usepackage{amsfonts}

\newcommand{\dist}[1]{\setlength{\abovecaptionskip}{#1cm}}
\newcommand{\disb}[1]{\setlength{\belowcaptionskip}{#1cm}}

\newcommand{\bolds}[1]{\boldsymbol{#1}}
\newcommand{\mathc}[1]{\mathcal{#1}}
\graphicspath{{images/}}

\usepackage[pagebackref=true,breaklinks=true,letterpaper=true,colorlinks,bookmarks=false]{hyperref}
\hypersetup{
    colorlinks=true,
    linkcolor=blue,
}

\iccvfinalcopy 

\ificcvfinal\fi

\begin{document}

\title{FakeMix Augmentation Improves Transparent Object Detection}

\author{
  Yang Cao$^1$ \quad
  Zhengqiang Zhang$^1$ \quad
  Enze Xie$^2$ \quad \\
  Qibin Hou$^3$ \quad 
  Kai Zhao$^4$ \quad 
  Xiangui Luo$^1$ \quad
  Jian Tuo$^1$ \quad \\
  Guangzhou Huya Information Technology Co.,Ltd$^1$ \\
  The University of Hong Kong$^2$   \\
  National University of Singapore$^3$ \quad Tencent Youtu$^4$ \\
  
}
\maketitle
\ificcvfinal\thispagestyle{empty}\fi

\begin{abstract}
Detecting transparent objects in natural scenes is challenging due to the low contrast in texture, brightness and colors. 
Recent deep-learning-based works reveal that it is effective to leverage boundaries for transparent object detection (TOD).
However, these methods usually encounter
boundary-related imbalance problem, leading to limited
generation capability.
%
%
%
Detailly, 
a kind of boundaries in the background, which share the same characteristics with boundaries of transparent objects
but have much smaller amounts, usually hurt the performance.
%
%
%
To conquer the boundary-related imbalance problem,
we propose a novel content-dependent data augmentation method
termed FakeMix.
Considering collecting these trouble-maker boundaries in the background
is hard without corresponding annotations, 
we elaborately generate them by appending the boundaries of transparent
objects from other samples into the current image during training,
which adjusts the data
space and improves the generalization
of the models.
Further, we present AdaptiveASPP, an enhanced version of ASPP, that
can capture multi-scale and cross-modality features dynamically.
%
Extensive experiments demonstrate that our methods clearly outperform the state-of-the-art methods.
We also show that our approach can also transfer well on related tasks,
in which the model meets similar troubles,
such as mirror detection, glass detection, and camouflaged object detection.
Code and models are available at: \href{https://github.com/yangcao1996/fanet}{github.com/yangcao1996/fanet}.
%

\end{abstract}

\section{Introduction}
\begin{figure}[htbp]
    \centering
    \begin{overpic}[width=1\columnwidth]{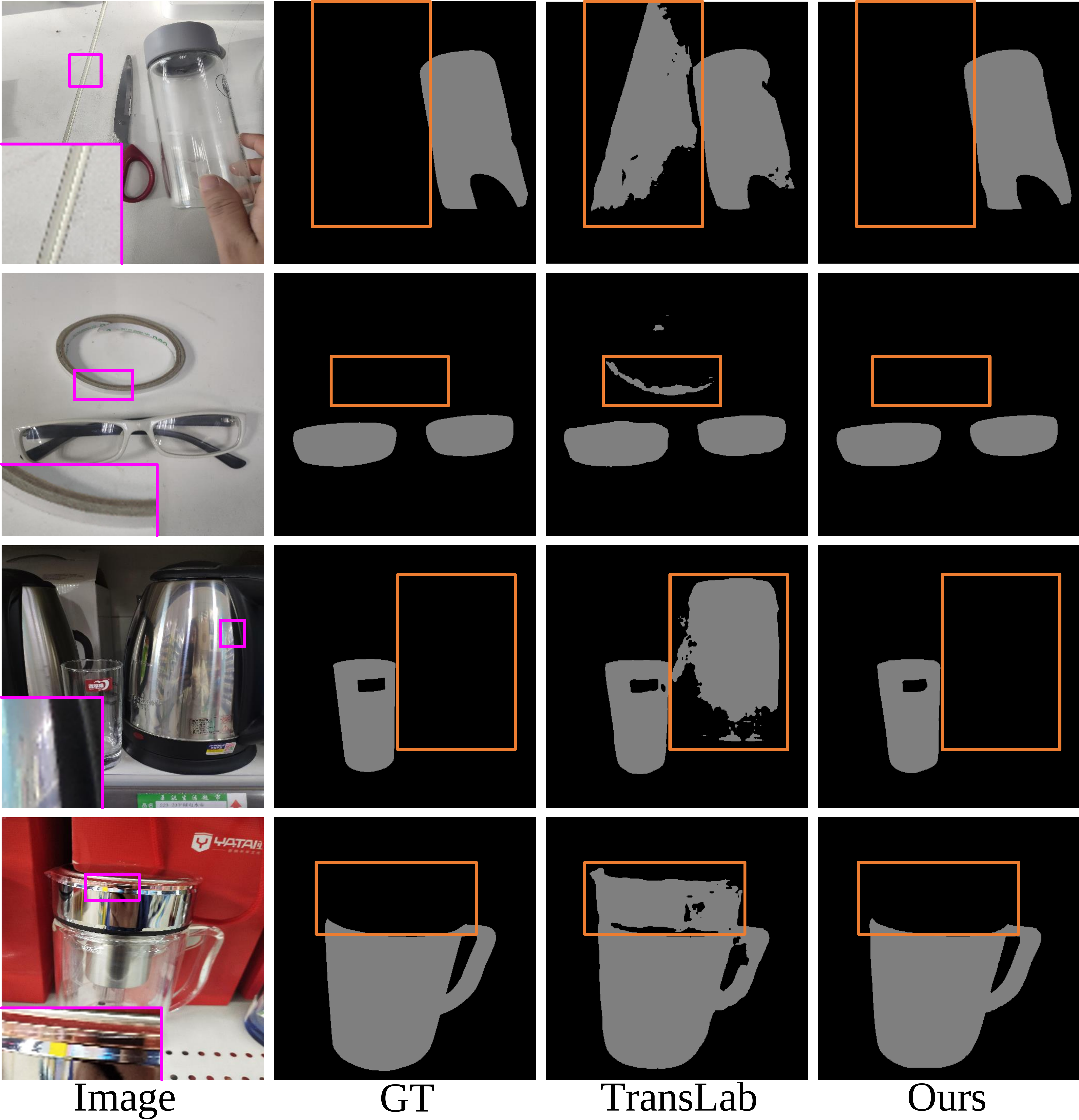}
    \end{overpic}
    \dist{-0.4}
    \disb{-0.5}
    \caption{Visual result comparisons between TransLab~\cite{xie2020segmenting}
    and our FANet. In the top two rows,
    the boundaries, labeled by the \textcolor{magenta}{pink} rectangle, have similar content
    on their two sides.
    In the bottom two rows, the
    boundaries, labeled by the \textcolor{magenta}{pink} rectangle, have obvious reflection and refraction.
    These boundaries are similar to the boundaries of transparent objects, leading to false detection labeled by the \textcolor{orange}{orange} rectangle.
    Our approach performs better
    due to FakeMix.}
    \label{fig:samples}
\end{figure}

Transparent Object Detection (TOD)~\cite{xie2020segmenting,xu2015transcut,chen2018tom} aims at detecting the transparent objects from the natural scene,
\textit{e.g.} windows, utensils, and glass doors, which widely exist in natural scenes. It is a relatively new and challenging task in the vision community.
Unlike opaque objects,
transparent objects often share similar textures,
brightness, and colors to the surrounding environments,
making them hard to detect for vision systems and even humans.
%
Because transparent objects are a widespread presence in our lives, 
they also have an influence on other tasks, such as depth estimation~\cite{silberman2012indoor},
segmentation~\cite{long2015fully}, and
saliency detection~\cite{hou2017deeply,fan2018salient}.
Therefore, how to accurately detect transparent objects is essential for many vision applications.

Benefiting from boundary clues, recent deep-learning-based work~\cite{xie2020segmenting}
have made great achievements in TOD.
However, deep-learning-based methods usually meet the data imbalance problem,
resulting in limited generalization.
%
When it comes to TOD, existing methods~\cite{xie2020segmenting} encounter
 serious boundary-related imbalance problem.
Specifically, the boundary-guided methods
pay too much attention to the Boundaries of Transparent objects (\textbf{T-Boundaries}).
Thus, they prefer to regard the background regions
surrounded by \textbf{Fake T-Boundaries} (sharing the same 
characteristics with T-boundaries but belonging to background)
as transparent objects.
For example, the boundaries labeled by the pink rectangle in the top two rows
of \figref{fig:samples} have similar content on their two sides.
In the bottom two rows of \figref{fig:samples},
the boundaries have obvious refraction or reflection.
These boundaries make their surroundings falsely predicted.
%
%
So how to distinguish T-Boundaries from Fake T-Boundaries is
key for addressing the boundary-related imbalance
problem in TOD. 

To improve the generation ability of models, 
some data augmentation methods
have been proposed in \cite{devries2017improved, zhang2017mixup, yun2019cutmix}.
Unfortunately, they are all content-agnostic and ignore the boundary clues,
leading to weak improvements for boundary-related imbalance problem.
In this paper, we propose a novel data augmentation method termed
\textbf{FakeMix}. FakeMix is content-dependent and could combat
the boundary-relate imbalance problem by balancing the data distribution of boundaries.
Concretely, we increase the proportion of Fake
T-Boundaries in the training set.
Notably, it is hard to collect Fake T-Boundaries directly from the background without corresponding annotations.
So we design a novel and efficient
method to generate Fake T-Boundaries.
Based on our observation,
Fake T-boundaries share the following
characteristic with T-boundaries:
(1) There are \textit{similar appearances on both sides} and
(2) There are \textit{obvious refraction or reflection}.
And the main difference between the two kinds
of boundaries is the appearances surrounded
by them.
Thus,
\emph{we generate Fake T-Boundaries by blending background
with the T-Boundaries}.
As the data distribution is balanced,
the model's capability of discriminating Fake T-Boundaries
and T-Boundaries can be improved.
Actually, as shown in \figref{fig:abla-features},
we find that FakeMix drives the model
to explore the apprearances inside, which are the key differences between the above two kinds of
boundaries.
%
%

Furthermore, we
improved ASPP~\cite{chen2017rethinking}
in an attention manner~\cite{hu2018squeeze}
and obtain the AdaptiveASPP module.
It inherits the characteristic 
of ASPP extracting multi-scale features,
but more importantly, it also benefits from
attention way, dynamically enhancing cross-modality
(segmentation and boundary)
features.
The exploration indicates adopting the multi-scale
and cross-modality features dynamically is
effective for TOD.
%

%
%
%
%
%

%

By adopting both FakeMix and AdaptiveASPP,
our FANet clearly outperforms
the state-of-the-art TOD and semantic segmentation methods.
Besides, we verify that our method also transfers well in relevant detection tasks,
such as mirror
detection~\cite{yang2019my,lin2020progressive},
glass detection~\cite{mei2020don}, and
camouflaged object detection~\cite{fan2020camouflaged}, showing the robustness of the proposed methods.
Extensive experiments on three corresponding real-world datasets
demonstrate that our FANet achieves state-of-the-art achievements.

In summary, our contributions are three-fold:
\vspace{-0.2cm}
\setlength{\parsep}{0cm}
\setlength{\topsep}{0cm}
\setlength{\itemsep}{0cm}
\begin{itemize}[leftmargin=*]
\setlength{\parsep}{0cm}
\setlength{\topsep}{0cm}
\setlength{\itemsep}{0cm}
\item We propose a novel content-dependent
data augmentation
method, called FakeMix, for transparent object detection.
It balances the data distribution of boundaries and weakens the boundary-relate imbalance problem in TOD.
In detail, we generate Fake T-Boundaries by blending background
with the T-Boundaries, which improve the model's
discrimination of the two kinds of boundaries by
scanning the appearances inside as shown in \figref{fig:abla-features}.
%

%
%
%
%
\item We improved  ASPP in 
an attention  manner  and 
 propose an AdaptiveASPP
to extract features adaptively in cross-modal
and multi-scale levels. Experiments help to validate its effectiveness.
%
\item Without bells and whistles,
      our model named FANet outperforms 
      state-of-the-art TOD methods.
      We further find more applications
      in related ``confusing region"
      detection tasks, i.e., mirror detection,
      glass detection and camouflaged object detection. FANet also gains state-of-the-art
      performances.
\end{itemize}
%
\begin{figure*}[t]
     \centering
    \begin{overpic}[width=1\textwidth]{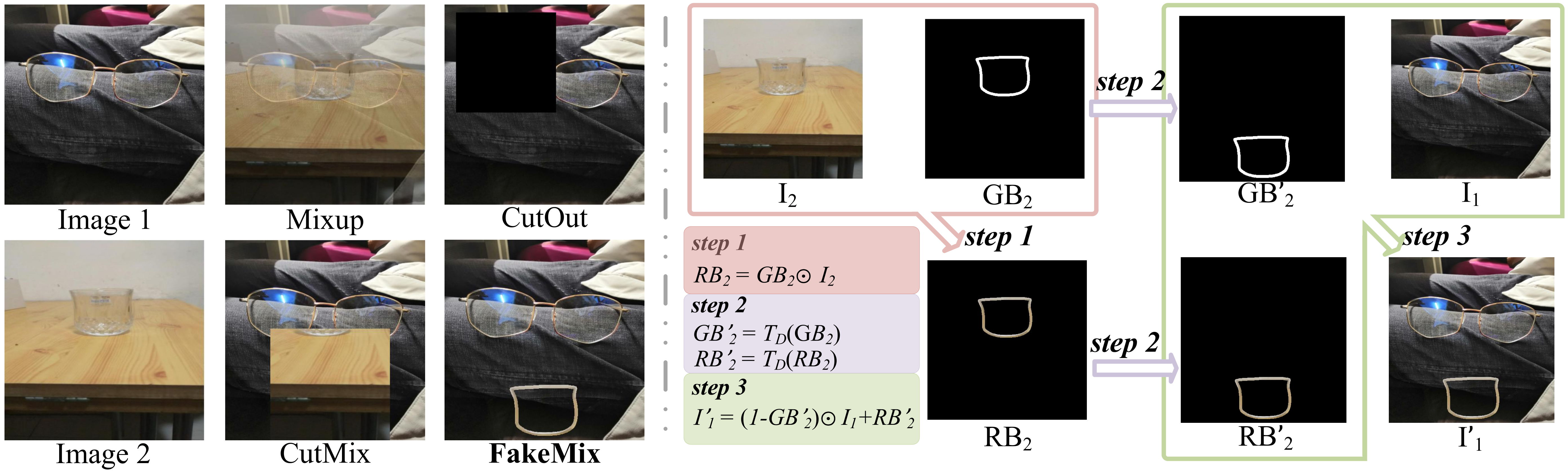}
    \end{overpic}
    \dist{-0.3}
    \caption{Comparison with other data augmentation methods and pipeline of FakeMix. Best view in color.
    $I_1$ is the image where we add the Fake T-Boundaries.
    $I_2$ is the image where we extract the T-Boundaries.
    $GB_2$ denotes the boundary label of $I_2$.
    Firstly, as ``\textit{\textbf{step 1}}",
    we extract the T-Boundaries from $I_2$, which is labeled
    $RB_2$. Secondly, we translate $GB_2$ and $RB_2$ randomly to
    the same position as ``\textit{\textbf{step 2}}".
    Then we get $GB_2^{'}$ and $RB_2^{'}$.
    Finally, we combine $GB_2^{'}$, $RB_2^{'}$ and $I_1$
    to get $I_1{'}$ as ``\textit{\textbf{step 3}}".
   }
    \label{fig:fboundary}
\end{figure*}
\section{Related Work}

\textbf{Data augmentation.}
To improve the generation and prevent the models from focusing too much on some regions on input
image,
some data augmentation methods\cite{devries2017improved, zhang2017mixup, yun2019cutmix} have been
proposed.
As shown in \figref{fig:fboundary},
Mixup~\cite{zhang2017mixup} combines two
images by linear interpolation.
Cutout~\cite{devries2017improved}
randomly removes some regions of the input
image.
CutMix~\cite{yun2019cutmix}
randomly replaces some regions with
a patch from another image.
These methods are simple and effective.
However, all of them are content-agnostic and ignore the boundary clues, resulting in limited improvements for TOD.
FakeMix combats the boundary-relate imbalance problem by
adjusting the data distribution of boundaries.

\textbf{Transparent object detection.}
Early work~\cite{xu2015transcut} proposed a model based on
LF-linearity and occlusion detection from the 4D light-field image.
\cite{chen2018tom} treats transparent object matting as
the refractive flow estimation problem.
Recently,~\cite{xie2020segmenting} proposed
a large-scale dataset for TOD, which consists
of 10428 images. Besides, they also designed
a boundary-aware segmentation method, named TransLab.
TransLab adopts the boundary clues to improve
the segmentation of transparent regions.
Our method, named FANet, also follows the
boundary-aware way.
While we found hidden troubles of boundary-aware methods:
some boundaries, which are similar to the boundaries of transparent objects, may hurt the detection.
Then we propose a novel
data augmentation method called FakeMix.
Besides, rather than directly introducing
ASPP in~\cite{xie2020segmenting}, we design an
AdaptiveASPP to extract features adaptively
for the segmentation and boundary branches respectively.
Besides, some topics that focus on
specific region detection are proposed recently,
such as \textbf{mirror detection}~\cite{yang2019my},
\textbf{glass detection}~\cite{mei2020don}
and \textbf{camouflaged object detection}~\cite{fan2020camouflaged}.
Considering boundary clues are also important for
distinguish the mirror, glass and camouflaged object,
these tasks might meet the similar problem with TOD.
We apply our FANet in the three
tasks above to measure the potential of our method
from more perspectives.
%
%

%
%

%
\section{Methods}
Facing the challenge of Transparent Object Detection (TOD), we propose  a 
method named FANet, which contains FakeMix and AdaptiveASPP.
The data augmentation method
named \textit{FakeMix} is proposed
to inspire the model to
exploit appearance clues, which will
be introduced in \secref{sec:fakeb}.
\textit{AdaptiveASPP}
module is designed to capture features of multiple fields-of-view adaptively
for the segmentation branch and boundary branch, respectively.
The details will be formulated in \secref{sec:aaspp}.

\subsection{FakeMix} \label{sec:fakeb}
TransLab~\cite{xie2020segmenting} proposes to 
exploit boundary clues to improve transparent detection.
However, we observe that some boundaries in the background may hurt the performance,
including
(1) the boundaries with \textit{similar contents on both sides} and
(2) the boundaries with \textit{obvious refraction or reflection}.
We define these boundaries as \textbf{Fake T-Boundaries}.
Fake T-Boundaries share the same characteristics with the boundary of the transparent object (\textbf{T-Boundaries}) but 
have a much smaller amount in the natural world.
%
%
Due to this boundary-related imbalance problem, the model prefers to regard the background regions surrounded by Fake T-Boundaries as transparent regions.
Thus, we propose a novel content-dependent data augmentation methods,
named FakeMix.
%
%
%
Considering that it is hard to collect Fake T-boundaries without corresponding annotations,
we elaborately manufacture them by appending T-Boundaries from other samples into the current image during training.
As the data distribution is balanced,
the discrimination ability of the model
between Fake T-Boundaries and T-Boundaries is improved during training.
%

\begin{figure*}[htbp]
    \centering
    \begin{overpic}[width=1\textwidth]{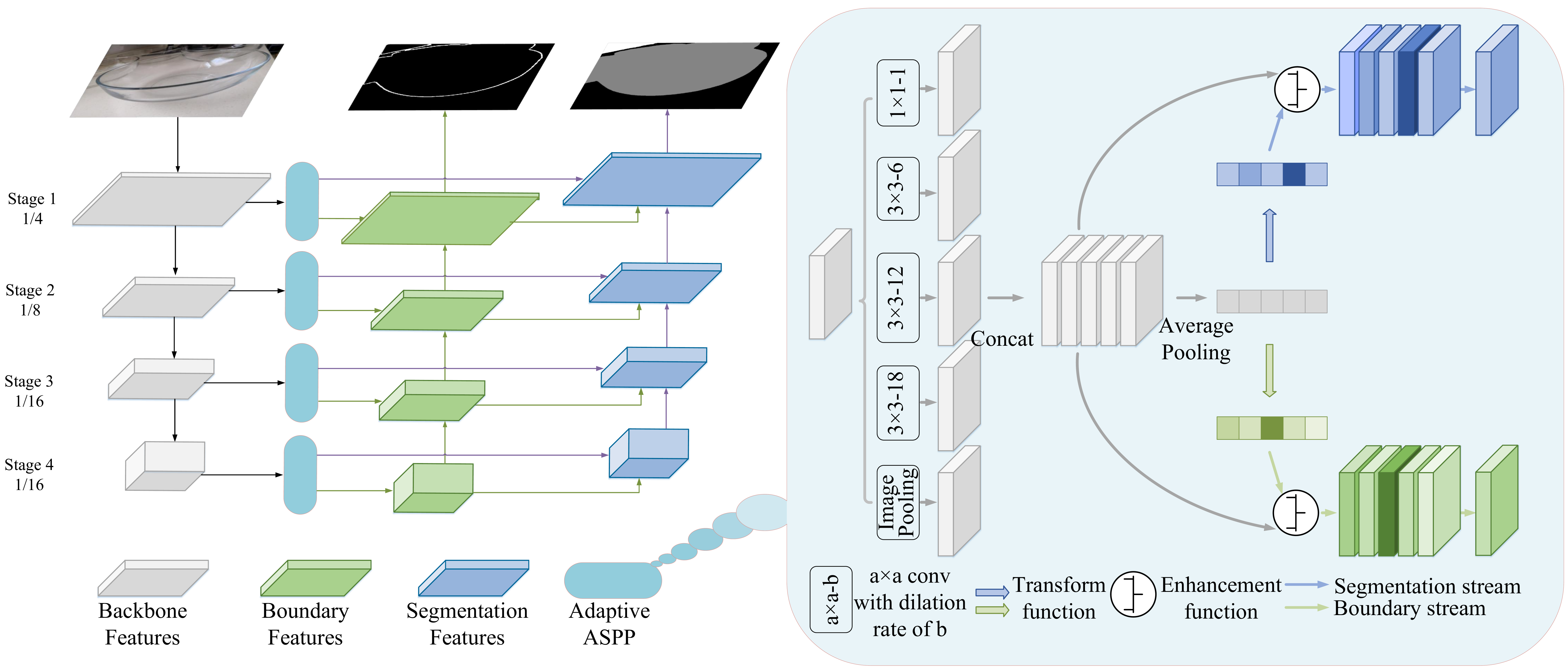}
    \end{overpic}
    \dist{-0.05}
    \disb{-0.4}    
    \caption{Overview architecture of FANet.
    The AdaptiveASPP modules are plugged at four stages of the backbone,
    which capture features of multiple fields-of-view adaptively
    for the segmentation branch and boundary branch, respectively. Then
    the features are integrated from bottom to top in each branch.
    In the segmentation branch, we follow \cite{xie2020segmenting} to
    fuse boundary features in the attention way.
    In the AdaptiveASPP. The transform function
    generates adaptive enhancement scores for the segmentation stream
    and boundary stream, respectively.
    Then the enhancement function enhances the features
    by enhancement scores adaptively for the two modalities.}
    \label{fig:archi}
\end{figure*}

Formally, let $I \in R^{W \times H \times C}$ denotes the input image
during training. $GS$ represents the segmentation label.
$GB$ is the boundary label generated from $GS$ as~\cite{xie2020segmenting}.
FakeMix combines the input image
$(I_1, GS_1, GB_1)$ with Fake T-Boundaries
from another training sample $(I_2, GS_2, GB_2)$
which is randomly selected.
Firstly, we extract the T-Boundaries
from $I_2$ as:
\begin{equation}
    RB_2 = GB_2 \odot I_2,
    \label{equ:tcontent}
\end{equation}
where $\odot$ denotes pixel-wise multiplication.
Then we apply the same affine transformation to
the boundary mask $GB_2$
and T-Boundaries $RB_2$,
which translates them to a random location.
The formulation can be written as:
\begin{equation}
    RB_2^{'} = T_D(RB_2),
    GB_2^{'} = T_D(GB_2),
    \label{equ:translate}
\end{equation}
where $T$ is the translation function.
$D$ denotes the translation vector:
\begin{equation}
    D = (\triangle x, \triangle y),
    \label{equ:tvector}
\end{equation}
$\triangle x$ and $\triangle y$
are uniformly sampled for every training sample
according to:
\begin{equation}
    \triangle x \sim \mathcal U(-\lambda w, \lambda w),
    \triangle y \sim \mathcal U(-\lambda h, \lambda h),
    \label{equ:tsample}
\end{equation}
where $w$ and $h$ are the width and height of the corresponding image. $\lambda$ is the parameter
to control the range of translation. We conduct ablation studies
of $\lambda$ in \secref{sec:abla:fboundary}.

According to \equref{equ:translate},
we get the randomly translated
boundary mask $GB_2^{'}$
and T-Boundaries $RB_2^{'}$.
Then we combine them with the input training sample
$I_1$ as:
\begin{equation}
    I_1^{'} = (1-GB_2^{'}) \odot I_1 + RB_2^{'},
    \label{equ:tcombine}
\end{equation}
Considering the T-Boundaries are separated away from the transparent appearances,
$I_1^{'}$ obtains more Fake T-Boundaries.

Further, we consider the choice of input sample 
as one bernoulli trial in each training iteration.
The trial results in one of two possible outcomes: either $I_1$ or $I_1^{'}$.
The probability mass function is:
\begin{equation}
    Pr(\hat{I}) =
    \begin{cases}
    p,     &\hat{I}=I, \\
    1-p, &\hat{I}=I_1^{'}. 
    \end{cases}
    \label{equ:bonl}
\end{equation}
\quad Finally, we get the training sample
$(\hat{I}, GS_1, GB_1)$ by the novel
FakeMix data augmentation method.
The pipeline can be visualized in \figref{fig:fboundary}.
\subsection{Architecture Overview} \label{sec:networks}
Following~\cite{xie2020segmenting},
ResNet50~~\cite{he2016deep} is used as the encoder
and a boundary branch is also included in the
decoder to help detect transparent regions.
An AdaptiveASPP is designed
to extract features of multiple fields-of-view
for both segmentation branch and boundary branch.

As shown in \figref{fig:archi},
our AdaptiveASPP
is plugged at four stages of
the encoder.
AdaptiveASPP extracts
features for the segmentation
and boundary branch, respectively.
Let $ \bolds{Z}^{(k, p)}$ denotes the features extracted by AdaptiveASPP
in the $p$-th ($p\in{[1, 4]}$) stage for branch $k$
($k\in{\{s, b}\}$, $s$ for the segmentation branch
and $b$ for the boundary branch).
Then the features are integrated from
bottom to top in each branch.
We formulate the features in the $p$-th stage
for branch $k$ of the decoder as $\bolds{M}^{(k,p)}$.
The cross-model feature fusion,
which is not the direction we delve,
simply follows~\cite{xie2020segmenting}
to apply boundary information in attention way
as:
\begin{small}
\begin{equation}
    \bolds{M}^{(s,p)} = 
    \begin{cases}
    \mathc{F}(\bolds{Z}^{(s,p)}+\bolds{Z}^{(s,p)} \odot \bolds{Z}^{(b,p)}), &p = 4, \\
    \\
    \mathc{F}(\bolds{Z}^{(s,p)}+\bolds{Z}^{(s,p)} \odot \bolds{Z}^{(b,p)} \\
    + \mathc{UP}(\bolds{M}^{(s,p+1)})), &p \in [1, 4),
    \end{cases}
    \label{equ:msp}
\end{equation}

\end{small}
where $\mathc{UP(\cdot)}$ means
the interpolation method which helps
to keep the same scale between features
from different stages. $\mathc{F}(\cdot)$ denotes a convolutional
function.
For the boundary branch, the integration
way is:
\begin{small}
\begin{equation}
    \bolds{M}^{(b,p)} =
    \begin{cases}
    \mathc{F}(\bolds{Z}^{(b,p)}), &p = 4, \\
    \mathc{F}(\bolds{Z}^{(b,p)}
    + \mathc{UP}(\bolds{M}^{(b,p+1)})), &p \in [1, 4).
    \end{cases}
\end{equation}
\end{small}
\quad The segmentation loss and boundary loss
supervise the two branches separately.
%

\subsection{AdaptiveASPP} \label{sec:aaspp}

To collect the information of multiple fields-of-view,
\cite{xie2020segmenting} adopts the Atrous Spatial Pyramid Pooling module (ASPP)~\cite{chen2017rethinking}. 
%
%
However, as exposed in~\cite{liu2020dynamic,zhao2019egnet},
detecting boundary and region focus on different targets and pay attention to different characteristics.
Thus we argue that richer features of multiple fields-of-view
with appropriate importances in cross-modal and multi-stage levels will exploit more promotion spaces.
Motivated by existing attention mechanism in~\cite{hu2018squeeze}, we carefully design an \textbf{AdaptiveASPP}
to capture features of multiple fields-of-view adaptively
for the boundary branch and segmentation branch.

As shown in \figref{fig:archi},
AdaptiveASPP firstly extracts features
of multiple fields-of-view by convolution kernels
with different dilation rates, which follows ASPP~\cite{chen2017rethinking} and can be formulated as: 
\begin{equation}
     \bolds{Y}_i = \mathc{F}_i(\bolds{X}),
    \label{equ:aspp}
\end{equation}

where $\bolds{X}$ means the input backbone features
of AdaptiveASPP.
$\mathc{F}_i(\cdot)$ denotes a convolutional
function and the subscript $i$ represents
the $i$-th dilation rate setting.
Let the feature maps extracted 
by different kernels be denoted by 
$\left\{\bolds{Y}_i | i\in[0, N)\right\}$,
where $N$ denotes the number of dilation rate settings.
In \figref{fig:archi}, we take $N=5$ for example.
Given $\left\{\bolds{Y}_i | i\in[0, N)\right\}$,
we concatenate them as $\bolds{Y}_N$
and conduct average pooling.
%
Then we have $N$-dimensional vector $\bolds{y}_N$ in which
$\bolds{y}_i$ is calculated as follows:
\begin{equation}
     \bolds{y}_i = AvePool(\bolds{Y}_i).
    \label{equ:vect}
\end{equation}
%

%
%
Then two \textbf{transform functions} for
the segmentation and boundary branch are
adopted to generate adaptive importances.
The formulation can be written as:
\begin{equation}
    \bolds{s}_N^k = \gamma{(\delta({\mathc{G}^k(\bolds{y}_N))})},       k\in\left\{s, b\right\},
    \label{equ:scores}
\end{equation}
where $k$ represents the modalities including boundary($b$)
and segmentation($s$). $\mathc{G}^k$ denotes the FC-ReLU-FC
block for corresponding modality. $\gamma(\cdot)$ indicates the normalization
function which maps the scores to $[0, 1]$. $\delta(\cdot)$ is the activation function.
The setting follows~\cite{li2020learning} as:
\begin{equation}
    \delta(\cdot) = max(Tanh(\cdot), 0).
    \label{equ:act}
\end{equation}

As we can see in \figref{fig:archi},
transform functions (refer to \equref{equ:scores})
generate adaptive importances for boundary
and segmentation modalities respectively.
Given importance vectors $\bolds{s}_N^k$,
we adopt the \textbf{enhancement function} as shown in \figref{fig:archi}
to enhance the features of multiple fields-of-view $\bolds{Y}_N$
for the two modalities. Then we can get
modality-specific features $\bolds{Z}_N^k$ as:
\begin{equation}
    \bolds{Z}_N^k = \bolds{Y}_N \times \bolds{s}_N^k + \bolds{Y}_N,
    k\in\left\{s, b\right\},
    \label{equ:enhance}
\end{equation}
In \equref{equ:enhance}, the residual connection is added
to preserve the original features $\bolds{Y}_N$ in
the enhancement function. Following the enhancement function, a convolutional block is
used to squeeze channel numbers.

\section{Experiments}
\subsection{Implementation Details}
The proposed model is implemented by
PyTorch~\cite{paszke2017automatic}.
In the encoder, we choose ResNet50~\cite{he2016deep}
to be the backbone as~\cite{xie2020segmenting}.
In the decoder, the channel numbers of convolutional
layers are set to 256.
The convolution type is set to separable convolution
~\cite{chen2018encoder} as~\cite{xie2020segmenting}.
The number of dilation rate is set to 7 experimentally
in AdaptiveASPP.
And the decoder
is randomly initialized.

\textbf{Training and Testing.}
During training, we
train our model for 400 epochs.
%
%
We choose the stochastic
gradient descent (SGD) optimizer.
The momentum and weight decay are set to
0.9 and 0.0005 respectively.
The learning rate is initialized to 0.01.
A poly strategy with the power of 0.9 is employed.
We use 8 V100 GPUs for our experiments.
Batch size is 4 on every GPU.
Random flip for the input image
is also conducted during training.
Following~\cite{xie2020segmenting},
we use dice loss~\cite{deng2018learning,milletari2016v,zhao2017pyramid}
for the boundary branch and CrossEntropy loss for the segmentation
branch.
Besides, images are resized to $512 \times 512$ during
training and testing as~\cite{xie2020segmenting}.
%
\subsection{Datasets}
We will introduce
the datasets adopted in our experiments.
The ablation experiments are conducted
on Trans10K, which is a challenging Transparent
Object Detection (TOD) dataset.
When comparing with state-of-the-art,
we find more applications of
our methods in related
topics for
specific region detection:
mirror detection~\cite{yang2019my}, glass detection~\cite{mei2020don}
and camouflaged object detection~\cite{fan2020camouflaged},
which could demonstrate the potential of our method. 
We keep exactly the same dataset and evaluation metrics
setting with the original paper.
%

\subsection{Ablation Study}

\subsubsection{Alternatives of FakeMix.}
We compare our FakeMix with
three different kinds of popular data augmentation method,
i.e., Mixup~\cite{zhang2017mixup}, Cutout~\cite{devries2017improved} and CutMix~\cite{yun2019cutmix}
on the latest
deep-learning-based TOD method TransLab~\cite{xie2020segmenting}.
According to \tabref{abla:aug_cmp},
our FakeMix gains the best performance in all
four metrics.
FakeMix is the only method that take boundaries
into consideration and combat the boundary-related
imbalance problem, leading to stable superiority.
%

%

\begin{table}[h]
    \centering
    \newcommand{\tf}[1]{\textbf{#1}}
    \small
    \begin{minipage}[htbp]{1\columnwidth}
    \centering
    \begin{tabular}{c|cccc}
    \hline
    \toprule
        Methods & Acc$\uparrow$ & mIoU$\uparrow$ & MAE$\downarrow$ & mBer$\downarrow$ \\
    \hline
    \toprule        
      Translab          & 83.04 & 72.10 & 0.166 & 13.30 \\
      Translab+Cutout   & 80.36 & 72.27 & 0.160 & 13.37 \\
      Translab+CutMix   & 81.32 & 72.21 & 0.162 & 12.94 \\
      Translab+Mixup    & 77.06 & 72.33 & 0.158 & 14.63 \\
      \hline
      Translab+FakeMix  & \tf{83.07} & \tf{73.01} & \tf{0.152} & \tf{12.76} \\
    \bottomrule
    \end{tabular}
    \dist{0.0}
    \disb{0.2}
    \caption{Comparison among different data augmentation methods.}
    \label{abla:aug_cmp}
    \end{minipage}
    \begin{minipage}[htbp]{1\columnwidth}
    \centering
    \begin{tabular}{c|cccc}
    \hline
    \toprule
        Methods & Acc$\uparrow$ & mIoU$\uparrow$ & MAE$\downarrow$ & mBer$\downarrow$ \\
    \hline
    \toprule        
        Convs       &79.32 &69.87 &0.184 &13.42 \\ 
        ASPPs       &82.17 &70.56 &0.170 &13.74 \\ 
      AdaptiveASPP  &\tf{85.81} &\tf{76.54} &\tf{0.139} &\tf{10.59} \\ 
    \bottomrule
    \end{tabular}
    \dist{0.0}
    \disb{0.2}
    \caption{Ablation for the alternative methods of AdaptiveASPP.
    ``Convs" denotes we adopt convolutional layers to replace
    AdaptiveASPP. ``ASPP" represents replacing AdaptiveASPP with ASPP.}
    \label{abla:alter}
    \end{minipage}

\end{table}
%
\subsubsection{AdaptiveASPP} \label{sec:abla:aaspp}

As described in \secref{sec:aaspp},
AdaptiveASPP could enhance features adaptively
in multiple fields-of-view and double-modality levels.
Here we delve into the effectiveness and details of AdaptiveASPP.
Specifically, we study (1) the effect of AdaptiveASPP,
(2) positions to adopt AdaptiveASPP,
(3) the activation function and (4) other specific settings.
Due to space constraints, the last three experiments are presented in
\textit{the supplementary materials}.
%
%
%
%
%

\textbf{Effectiveness of AdaptiveASPP}.
To explore the effect of AdaptiveASPP,
we replace the AdaptiveASPP in \figref{fig:archi}
with two convolutional layers or two ASPPs~\cite{chen2017rethinking}
to generate features for the segmentation branch
and the boundary branch separately.
As we can
see in \tabref{abla:alter}, the
performance elaborates on the superiority
of AdaptiveASPP.
Compared with Convs and ASPPs which
treat the features with the same importance,
our AdaptiveASPP
enhances features adaptively,
leading to better performances.
\begin{figure}[t]
    \centering
    \begin{overpic}[width=1\columnwidth]{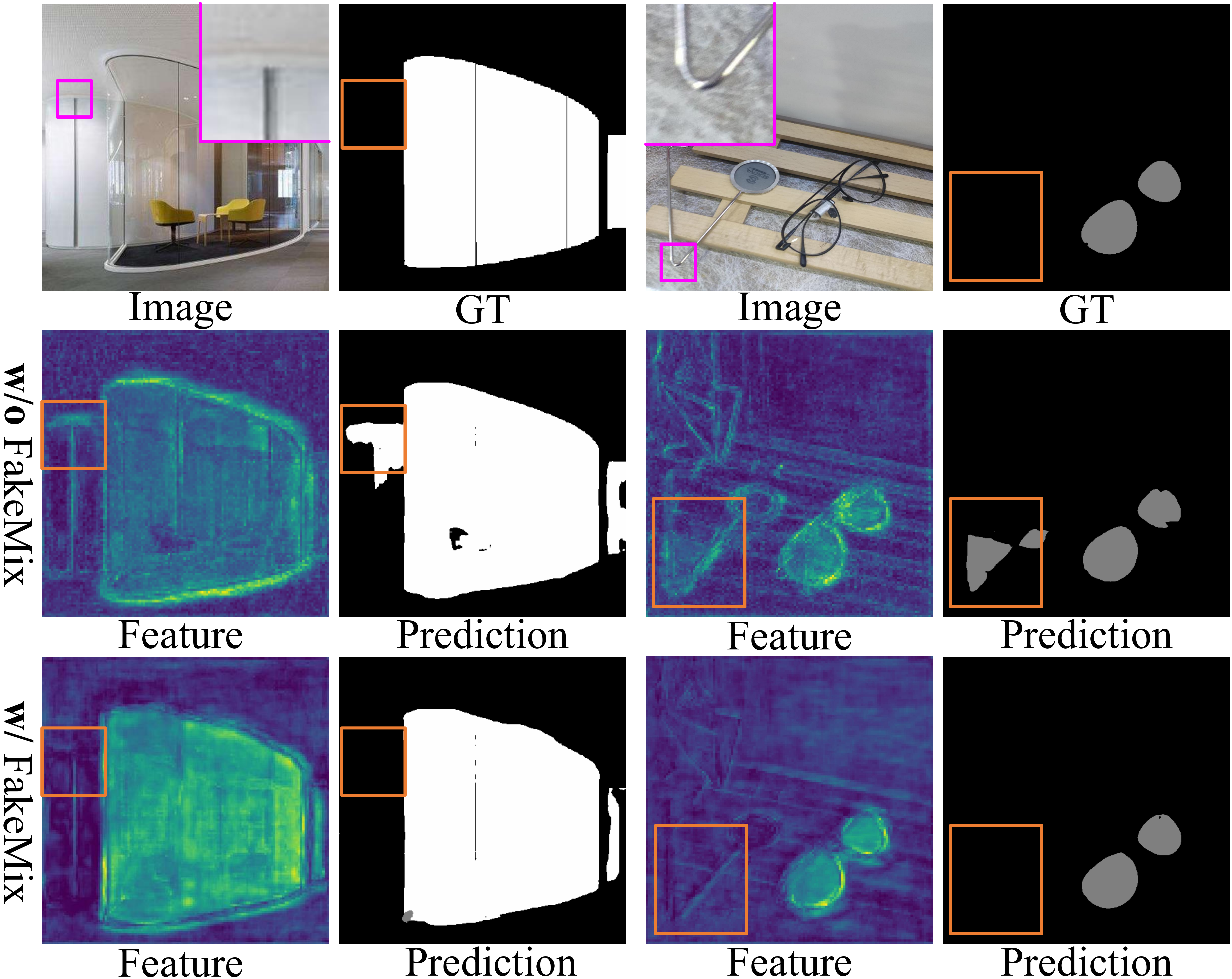}
    \end{overpic}
    \dist{-0.3}
     \disb{-0.4}       
    \caption
    {Comparison of features w/o FakeMix
    and w/ FakeMix. Best view in color and zoom-in.
    Features and predictions in the 2nd row labeled ``w/o FakeMix" tell that the model
    without utilizing FakeMix is confused by the boundaries labeled by the
    \textcolor{magenta}{pink} rectangle,
    leading to failed detection labeled by the \textcolor{orange}{orange} rectangle.
    Features and predictions in the 3rd row labeled ``w/ FakeMix" show that the model focuses
    on the appearances of transparent objects,
    which are the key differences between T-Boundaries
    and Fake T-Boundaries,
    resulting in better prediction results labeled by the \textcolor{orange}{orange} rectangle.} 
    \label{fig:abla-features}
\end{figure}

\subsubsection{Delving into FakeMix.} \label{sec:abla:fboundary}
\quad As demonstrated in \secref{sec:fakeb},
FakeMix enhances the discrimination ability of the model during training.
Here we study the effectiveness and different settings
of FakeMix.
Specifically, we firstly visualize (1) the features
of our model trained w/o and w/ FakeMix
to observe how FakeMix works on the model.
Then we conduct ablation experiments on different settings,
including
(2) the range of translation, (3) the probability
 of adding Fake T-Boundaries,
(4) the content of Fake T-Boundaries
and (5) the number of Fake T-Boundaries.

\textbf{Look deeper into the features.}
To study how FakeMix works,
we look deeper into FANet and visualize the features following
the samilar way with \cite{liu2020dynamic}.
In detail, the features visualized are from the first stage of
the decoder, namely $M^{s, 1}$ in \equref{equ:msp} in the paper.
The max function is applied along the channel dimension to
get the visualization features.
In \figref{fig:abla-features},
features and predictions in the 2nd row
are from FANet trained without FakeMix.
As we can see, these features notice
the Fake T-Boundaries obviously,
which resulting in failed prediction
in nearby regions.
The features and predictions in the 3rd row
are from FANet trained with FakeMix.
These features 
pay more attention to transparent
appearances, which are the key differences between T-Boundaries and Fake T-Boundaries,
leading to better predictions.
Research on human
perception of transparent
objects~\cite{schluter2019visual} elaborates that some optical
phenomena in transparent
appearances, e.g., the refraction and reflection, may bring potential
clues, which might be utilized by our FANet trained with FakeMix.
%

%
%

\begin{table}[t]
    \centering
    \newcommand{\tf}[1]{\textbf{#1}}
    \small
    \begin{minipage}[htbp]{1\columnwidth}
    \centering
    \begin{tabular}{c|cccc}
    \hline
    \toprule
        $\lambda$ & Acc$\uparrow$ & mIoU$\uparrow$ & MAE$\downarrow$ & mBer$\downarrow$ \\
    \hline
    \toprule
        -        &85.81 &76.54 &0.139 &10.59 \\ 
        0        &87.23 &76.99 &0.129 &10.22 \\ 
        1/3        &86.99 &76.89 &0.130 &10.32 \\ 
        1/2        &\tf{87.36} &\tf{77.62} &\tf{0.128} &\tf{9.93}  \\ 
        2/3        &87.14 &77.54 &0.129 &10.20  \\ 
    \bottomrule
    \end{tabular}
    \dist{0.0}
    \disb{0.2}          
    \caption{Ablation for the range of translation, namely $\lambda$ in \equref{equ:tsample}. ``-"
represents that the model is trained without FakeMix.}
    \label{tab:fdis}    
    \end{minipage}
    
    \begin{minipage}[htbp]{1\columnwidth}
    \centering
    \begin{tabular}{c|cccc}
    \hline
    \toprule
        $p$ & Acc$\uparrow$ & mIoU$\uparrow$ & MAE$\downarrow$ & mBer$\downarrow$ \\
    \hline
    \toprule
        0        &85.81 &76.54 &0.139 &10.59 \\ 
        1/3        &86.93 &77.60 &\tf{0.126} &10.03 \\ 
        1/2        &\tf{87.36} &\tf{77.62} &0.128 &\tf{9.93}  \\ 
        2/3        &86.98 &77.40 &0.129 &10.17  \\ 
    \bottomrule
    \end{tabular}
    \dist{0.0}
    \disb{0.2}          
    \caption{Ablation for the probability of adding Fake T-Boundaries, namely $p$ in \equref{equ:bonl}. ``0"
represents that the model is trained without FakeMix.}
    \label{tab:fprob}    
    \end{minipage}
    
    \begin{minipage}[htbp]{1\columnwidth}
    \centering
    \begin{tabular}{c|cccc}
    \hline
    \toprule
        $RB_2$ & Acc$\uparrow$ & mIoU$\uparrow$ & MAE$\downarrow$ & mBer$\downarrow$ \\
    \hline
    \toprule
        -        &85.81 &76.54 &0.139 &10.59 \\ 
        zero        &86.80 &76.71 &0.130 &10.21 \\ 
        mean        &86.75 &77.08 &0.132 &10.23 \\ 
        random        &86.22 &76.62 &0.132 &10.33  \\ 
        boundary        &\tf{87.36} &\tf{77.62} &\tf{0.128} &\tf{9.93}  \\ 
    \bottomrule
    \end{tabular}
    \dist{0.0}
    \caption{Ablation for the content of Fake T-Boundaries, namely $RB_2$ in \equref{equ:tcontent}. ``-"
represents that the model is trained without FakeMix. ``zero" denotes
that the values of Fake T-Boundaries are set to 0.
``mean" denotes that the values of Fake T-Boundaries are set to the mean value of Trans10K. ``random" represents
that the values of Fake T-Boundaries are set to the random
region of images. ``boundary" means the values
are set to the content of the boundary from
transparent objects as computed in \equref{equ:tcontent}.}
    \label{tab:fcontent}    
    \end{minipage}    

    \begin{minipage}[htbp]{1\columnwidth}
    \centering
    \begin{tabular}{c|cccc}
    \hline
    \toprule
        Numbers & Acc$\uparrow$ & mIoU$\uparrow$ & MAE$\downarrow$ & mBer$\downarrow$ \\
    \hline
    \toprule
        0        &85.81 &76.54 &0.139 &10.59 \\ 
        1        &87.12 &\tf{78.00} &0.127 &9.96 \\ 
        2        &87.14 &77.71 &\tf{0.126} &9.99 \\ 
        3        &\tf{87.36} &77.62 &0.128 &\tf{9.93}  \\ 
        4        &86.61 &77.87 &0.127 &10.21 \\ 
    \bottomrule
    \end{tabular}
    \vspace{-0.3cm}    
    \caption{Ablation for the number of Fake T-Boundaries
    we put on the input image. To gain ``n" Fake T-Boundaries,
    we repeat the progress described in Sec.3.3 of our paper ``n" times.}
    \label{tab:fnum}
    \end{minipage}    
\end{table}

\textbf{Different settings.} We analyze
\textbf{the range of translation},
namely $\lambda$ in \equref{equ:tsample}.
As shown in \tabref{tab:fdis},
different settings from the second
row to the fourth row
all bring improvements
compared with the one without FakeMix,
which validates the
effectiveness and practicability of our FakeMix.
Experimentally, we set $\lambda$ to $1/2$.
\begin{table*}[t]
    \centering
    \small
    \newcommand{\mc}{\multicolumn}
    \newcommand{\mr}{\multirow}
    \newcommand{\tr}[1]{\textcolor{red}{#1}}
    \newcommand{\tg}[1]{\textcolor{green}{#1}}
    \newcommand{\tb}[1]{\textcolor{black}{#1}}
    \newcommand{\tc}[1]{\textcolor{black}{#1}}
    \newcommand{\tf}[1]{\textbf{#1}}
    \begin{tabular}{c|c|c|cc|cc|cc}
    \hline
    \toprule
        \mr{2}*{Method} & \mr{2}*{ACC $\uparrow$} & \mr{2}*{MAE $\downarrow$}
        & \mc{2}{c|}{IoU $\uparrow$} & \mc{2}{c|}{BER $\downarrow$}
        & \mc{2}{c}{Computation} \\ 
        \cline{4-9}
               &     &     &  Stuff & Things & Stuff & Things & Params/M & FLOPs/G \\
    \midrule[0.5pt]
        ICNet~\cite{zhao2018icnet} & 52.65 & 0.244 & 47.38 & 53.90 & 29.46 & 19.78 & 8.46 & 10.66 \\
        BiSeNet~\cite{yu2018bisenet} & 77.92 & 0.140 & 70.46 & 77.39 & 17.04 & 10.86 & 13.30 & 19.95 \\
        DenseASPP~\cite{yang2018denseaspp} & 81.22 & 0.114 & 74.41 & 81.79 & 15.31 & 9.07 & 29.09 & 36.31 \\
        FCN~\cite{long2015fully} & 83.79 & 0.108 & 74.92 & 84.40 & 13.36 & 7.30 & 34.99 & 42.35 \\
        UNet\cite{ronneberger2015u} & 51.07 & 0.234 & 52.96 & 54.99 & 25.69 & 27.04 & 13.39 & 124.62 \\
        OCNet~\cite{yuan2018ocnet} & 80.85 & 0.122 & 73.15 & 80.55 & 16.38 & 8.91 & 35.91 & 43.43 \\
        DUNet~\cite{jin2019dunet} & 77.84 & 0.140 & 69.00 & 79.10 & 15.84 & 10.53 & 31.21 & 123.35 \\
        PSPNet~\cite{zhao2017pyramid} & 86.25 & 0.093 & 78.42 & 86.13 & 12.75 & 6.68 & 50.99 & 187.27 \\
        DeepLabv3+~\cite{chen2018encoder}  & 89.54 & 0.081 & 81.16 & 87.90 & 10.25 & 5.31 & 28.74 & 37.98 \\ \midrule[0.5pt]
        TransLab~\cite{xie2020segmenting} & 92.69 & 0.063 & 84.39 & 90.87 & 7.28 & 3.63 & 40.15 & 61.27 \\
        TransLab~\cite{xie2020segmenting} + FakeMix & 93.14 & 0.057 & 85.62 & 91.91 & 6.68 & 3.28 & 40.15 & 61.27 \\ \midrule[0.5pt]
        FANet & \tf{94.00} &\tf{0.052} & \tf{87.01} & \tf{92.75} & \tf{6.08} & \tf{2.65} & 35.39 & 77.57 \\ 
        FANet + FakeMix & \tf{94.93} &\tf{0.046} & \tf{88.29} & \tf{93.42} & \tf{5.43} & \tf{2.36} & 35.39 & 77.57 \\ 
    \bottomrule
    \end{tabular}
    \dist{-0.05}
    \caption{Comparison between stuff set and thing set of Trans10K. Note that
FLOPs is computed with one $512 \times 512$ image.}
    \label{tab:sandt}
\end{table*}
\begin{table*}[t]
    \centering
    \small
    \newcommand{\mc}{\multicolumn}
    \newcommand{\mr}{\multirow}
    \newcommand{\tr}[1]{\textcolor{red}{#1}}
    \newcommand{\tg}[1]{\textcolor{green}{#1}}   
    \newcommand{\tb}[1]{\textcolor{black}{#1}}
    \newcommand{\tc}[1]{\textcolor{black}{#1}}
    \newcommand{\tf}[1]{\textbf{#1}}
    
    \begin{tabular}{c|ccc|ccc|ccc|ccc}
    \hline
    \toprule
        \mr{2}*{Method} & \mc{3}{c|}{mIoU $\uparrow$} & \mc{3}{c|}{Acc $\uparrow$}
        & \mc{3}{c|}{MAE $\downarrow$} & \mc{3}{c}{mBER $\downarrow$} \\ 
        \cline{2-13}
               & Hard & Easy & All & Hard & Easy & All & Hard & Easy & All & Hard & Easy & All \\
    \hline
    \toprule
        ICNet~\cite{zhao2018icnet}  & 33.44 & 55.48 & 50.65 & 35.01 & 58.31 & 52.65 & 0.408 & 0.200 & 0.244 & 35.24 & 21.71 & 24.63 \\
        BiSeNet~\cite{yu2018bisenet}             & 56.37 & 78.74 & 73.93 & 62.72 & 82.79 & 77.92 & 0.282 & 0.102 & 0.140 & 24.85 & 10.83 & 13.96 \\
        DenseASPP~\cite{yang2018denseaspp}                    & 60.38 & 83.11 & 78.11 & 66.55 & 86.25 & 81.22 & 0.247 & 0.078 & 0.114 & 23.71 & 8.85 & 12.19 \\
        FCN~\cite{long2015fully}             & 62.51 & 84.53 & 79.67 & 68.93 & 88.55 & 83.79 & 0.239 & 0.073 & 0.108 & 20.47 & 7.36 & 10.33 \\
        UNet\cite{ronneberger2015u} &37.08 & 58.60 & 53.98 & 37.44 & 55.44 & 51.07 & 0.398 & 0.191 & 0.234 & 36.80 & 23.40 & 26.37 \\
        OCNet~\cite{yuan2018ocnet}                    & 59.75 & 81.53 & 76.85 & 65.96 & 85.63 & 80.85 & 0.253 & 0.087 & 0.122 & 23.69 & 9.43 & 12.65 \\
        DUNet~\cite{jin2019dunet}             & 55.53 & 79.19 & 74.06 & 60.50 & 83.41 & 77.84 & 0.289 & 0.100 & 0.140 & 25.01 & 9.93 & 13.19 \\
        PSPNet~\cite{zhao2017pyramid}                    & 66.35 & 86.79 & 82.38 & 73.28 & 90.41 & 86.25 & 0.211 & 0.062 & 0.093 & 20.08 & 6.67 & 9.72 \\   
        DeepLabv3+~\cite{chen2018encoder}                 & 69.04 & 89.09 & 84.54 & 78.07 & 93.22 & 89.54 & 0.194 & 0.050 & 0.081 & 17.27 & 4.91 & 7.78 \\       
         \midrule[0.5pt]        
        TransLab~\cite{xie2020segmenting}                                               & 72.10 & 92.23 & 87.63 & 83.04 & 95.77 & 92.69 & 0.166 & 0.036 & 0.063 & 13.30 & 3.12 & 5.46 \\    
        
        TransLab~\cite{xie2020segmenting} + FakeMix                                              & 73.01 & 93.19 & 88.76 & 83.07 & 96.37 & 93.14 & 0.152 & 0.032 & 0.057 & 12.76 & 2.71 & 4.98 \\     
         \midrule[0.5pt]        
         FANet                                               & \tf{76.54} & \tf{93.77} & \tf{89.88} & \tf{85.81} & \tf{96.62} & \tf{94.00} & \tf{0.139} & \tf{0.029} & \tf{0.052} & \tf{10.59} & \tf{2.43} & \tf{4.37} \\    
         FANet + FakeMix                                              & \tf{77.62} & \tf{94.70} & \tf{90.86} & \tf{87.36} & \tf{97.36} & \tf{94.93} & \tf{0.128} & \tf{0.024} & \tf{0.046} & \tf{9.93} & \tf{2.03} & \tf{3.89} \\

    \bottomrule
    \end{tabular}
    \dist{-0.05}    
    \disb{-0.5}   
    \caption{Comparison between hard set and easy set of Trans10K.}
    \label{tab:hande}
\end{table*}
%
%
%
Besides, we study \textbf{the probabilities of adding Fake T-Boundaries}, i.e., the $p$ in \equref{equ:bonl}.
Results from \tabref{tab:fprob} demonstrates that using non-zero probability settings could boost  performance.
We choose $p=1/2$.
%
%

\textbf{The content of Fake T-Boundaries.}
Furthermore, we explore the content of Fake T-Boundaries.
As shown in \tabref{tab:fcontent},
the boundary of transparent objects
can provide the best performance.
This is because the boundaries of transparent objects
have two characteristics:
(1) the boundaries with similar appearances on their two sides;
(2) the boundaries with obvious refraction or reflection.
These boundaries are most likely to cause failed detection.
Considering choosing the boundaries of transparent objects as the content of Fake T-Boundaries
will help models to distinguish these boundaries, the ``boundary" in \tabref{tab:fcontent}
achieves better performance.

\textbf{The number of Fake T-Boundaries.}
we try different numbers of Fake T-Boundaries in FakeMix.
As shown in \tabref{tab:fnum},
we found that the performances,
when the number is greater than 0,
become generally 
better than the model trained without
without FakeMix.

\begin{figure}[t]
    \centering
    \begin{overpic}[width=1\columnwidth]{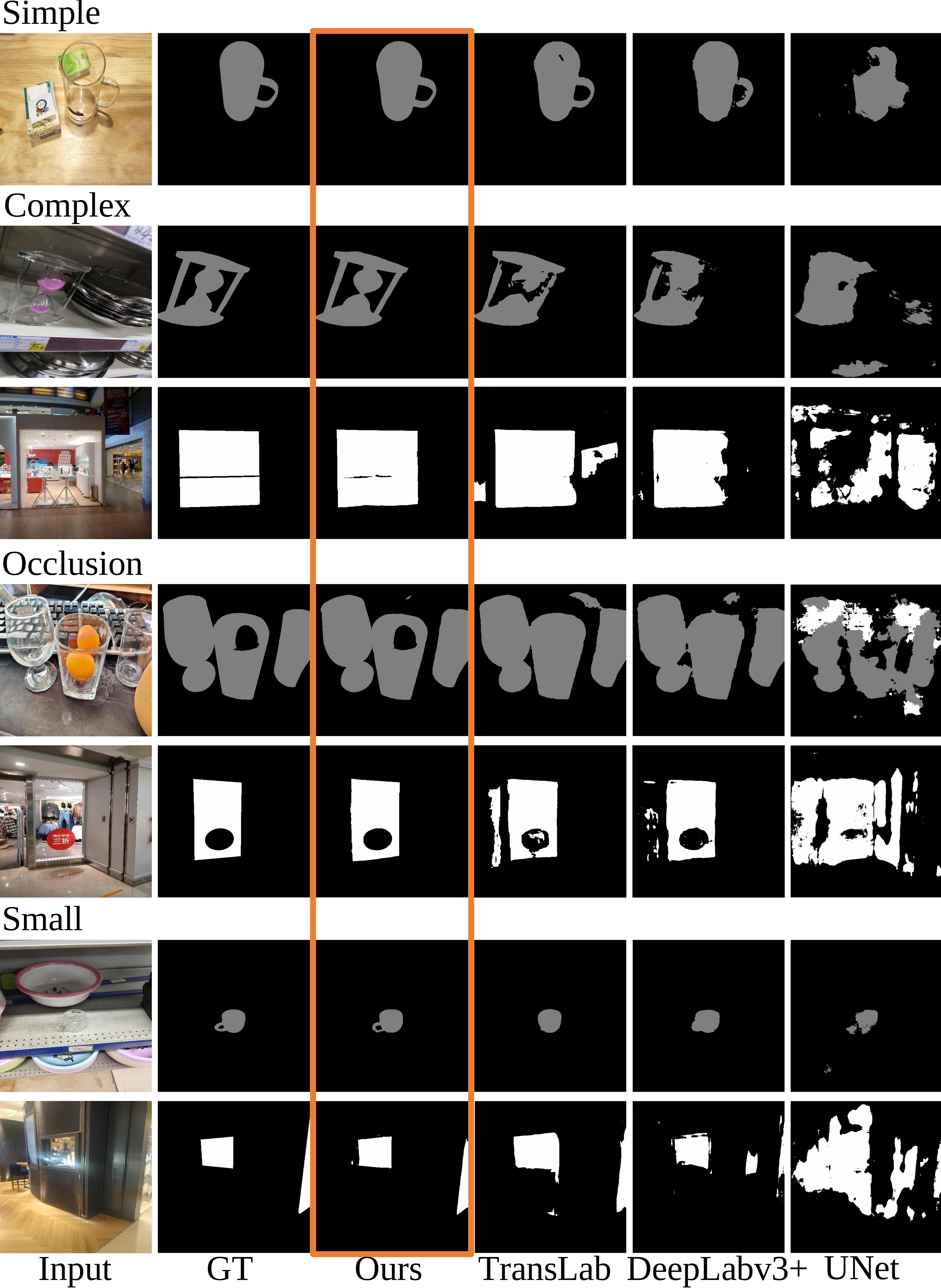}
    \end{overpic}
    \vspace{-0.8cm}
    \caption
    {Visual comparison on Trans10K~\cite{xie2020segmenting}.}
    \label{fig:viscmp}
\end{figure}

\subsection{Compare with the State-of-the-art.}
This section compares our FANet with alternative
methods on the large-scale \textbf{transparent
object detection} dataset Trans10K. 
Besides,
we apply our FANet in related confusing region
detection tasks, i.e., mirror detection,
glass detection and camouflaged object detection
to measure the potential
of our method in more topics. Noted that
we retrain and evaluate our model respectively
following the same
train/test setting with the original papers
for the fair comparison.
The comparison on \textbf{mirror detection},
\textbf{glass detection} and \textbf{camouflaged object detection}
could be found in \textit{the supplementary materials}.

%
%
We compare our FANet with state-of-the-art
methods named TransLab~\cite{xie2020segmenting}
and main-stream semantic segmentation methods
on TOD dataset Trans10K.
\tabref{tab:sandt} reports the quantitative results
of four metrics in both easy/hard set.
\tabref{tab:hande} reports the quantitative results
in things/stuff set.
As we can see, benefitting from our
FakeMix and AdaptiveASPP,
FANet outperform alternative methods
significantly.
Furthermore, we compare
the qualitative results
of TOD methods.
Especially, we summarize
several challenging scenes in TOD:
the complex scene, the scene with occlusion,
and the scene with small objects.
More hard scenes, i.e., the scene with multi-category
objects and the scene with unnoticeable objects,
are shown in \textit{the supplementary materials}.
As shown in \figref{fig:viscmp},
the 1st row shows a simple example
in which most methods perform well.
In the 2nd-3rd rows, we sample some images
in which the scenes are complex.
The scenes with occlusion
are shown in the 4th-5th rows.
As we can see, in complex and occlusion scenes,
our model avoids bad influences of the boundaries
from non-transparent regions and gain more complete
results with better details, benefiting from our FakeMix.
Then we show other challenging situations
in which the transparent objects are small.
As we can see in the last three rows of
\figref{fig:viscmp}, considering AdaptiveASPP
could capture features of multiple fields-of-view for
segmentation and boundary branches adaptively,
our model locates the small transparent objects
well and segments them finely.

\section{Conclusion}
In this paper, we proposed a novel content-dependent
data augmentation method termed FakeMix.
FakeMix weakens the boundary-related
imbalance problem in the natural word and
strengthens the discrimination ability of the model
for Fake T-Boundaries and T-Boundaries.
%
%
%
Besides,
we design an \textbf{AdaptiveASPP}
module to capture features of multiple fields-of-view
adaptively for the segmentation and boundary
branches, respectively.
%
%
Benefiting from FakeMix and
AdaptiveASPP, our FANet surpasses
state-of-the-art TOD methods significantly.
Further, FANet also gains state-of-the-art
performance in
related confusing region detection
tasks, i.e., mirror detection,
glass detection and camouflaged object detection.
%


\bibliographystyle{IEEEtran}
\bibliography{transseg}

\end{document}